\pdfoutput=1
\documentclass[Afour,sageh,times]{sagej}

\PassOptionsToPackage{numbers,sort&compress}{natbib}

\usepackage{url}
\usepackage[colorlinks,bookmarksopen,bookmarksnumbered,citecolor=blue,urlcolor=blue]{hyperref}
\usepackage{multirow}
\usepackage{booktabs}
\usepackage{xcolor}
\usepackage{capt-of}
\usepackage{nameref}
\usepackage[most]{tcolorbox}
\newtcolorbox{promptbox}[1]{
  enhanced,
  breakable,
  colback=gray!3,
  colframe=gray!45,
  colbacktitle=gray!18,
  coltitle=black,
  fonttitle=\bfseries\small,
  title=#1,
  boxrule=0.45pt,
  arc=1.5pt,
  left=5pt,
  right=5pt,
  top=4pt,
  bottom=4pt,
  toptitle=3pt,
  bottomtitle=3pt,
  before skip=6pt,
  after skip=6pt
}

\newcommand{\ours}{SciCore-Mol}
\bibpunct{[}{]}{,}{n}{}{,}

\begin{document}

\title{SciCore-Mol: Augmenting Large Language Models with Pluggable Molecular Cognition Modules}

\author{
Yuxuan Chen\textsuperscript{1},
Changwei Lv\textsuperscript{2},
Yunduo Xiao\textsuperscript{2},
Zhongjing Du\textsuperscript{1},
Daquan Zhou\textsuperscript{1},
Yukun Yan\textsuperscript{2},
Zheni Zeng\textsuperscript{3,*}
and Zhiyuan Liu\textsuperscript{2,*}
}

\affiliation{
\textsuperscript{1}School of Electronic and Computer Engineering, Peking University, Shenzhen, China\\
\textsuperscript{2}Tsinghua University, Beijing, China\\
\textsuperscript{3}School of Intelligence Science and Technology, Nanjing University, Suzhou, China
}

\corrauth{Zheni Zeng, Zhiyuan Liu}

\email{zengzn@nju.edu.cn, liuzy@tsinghua.edu.cn}

\begin{abstract}

Large Language Models (LLMs) are central to the one-for-all intelligent paradigm, but they face a fundamental challenge when dealing with heterogeneous scientific data such as molecules: the inherent gap between discrete linguistic symbols and topological molecular or continuous reaction data leads to significant information loss and semantic noise in text-based reasoning. We propose \textbf{SciCore-Mol}, a modular framework that bridges this gap through three deeply integrated pluggable cognitive modules: a topology-aware perception module, a latent diffusion-based molecular generation module, and a reaction-aware reasoning module. Each module is coupled to the LLM backbone through learned representation interfaces, enabling richer information exchange than is possible with text-only tool feedback. Our experiments on diverse chemical tasks demonstrate that \ours{} achieves strong comprehensive performance across molecular understanding, generation, reaction prediction, and general chemistry knowledge, with an 8B-parameter open-source system that is competitive with and in several dimensions surpasses proprietary large models. This work provides a systematic blueprint for equipping LLMs with scientific expertise through decoupled, pluggable, and flexibly orchestrated modules, with direct implications for drug design, chemical synthesis, and broader scientific discovery.

% Large Language Models (LLMs) are central to the one-for-all intelligent paradigm, but they struggle with heterogeneous scientific data such as molecules. The gap between discrete linguistic symbols and topological molecular structures or continuous reaction data can cause information loss and semantic noise in text-based reasoning. We propose \textbf{SciCore-Mol}, a modular framework that bridges this gap through three integrated pluggable cognitive modules: a topology-aware perception module, a latent diffusion-based molecular generation module, and a reaction-aware reasoning module. Each module is coupled to the LLM backbone through learned representation interfaces, enabling richer information exchange than text-only tool feedback. Experiments across diverse chemical tasks show that \ours{} achieves strong performance in molecular understanding, generation, reaction prediction, and general chemistry knowledge; the 8B-parameter open-source system is competitive with, and in several settings surpasses, proprietary large models. This work offers a blueprint for equipping LLMs with scientific expertise through decoupled, pluggable modules.

\end{abstract}

\maketitle
% sections/1_introduction.tex
\section{Introduction}\label{sec:introduction}

Large language models (LLMs) have demonstrated remarkable reasoning and knowledge storage capabilities, leading to their widespread application in fields such as biochemistry. However, applying LLMs to scientific data exposes a fundamental mismatch: models designed for discrete symbolic sequences must often process heterogeneous objects such as molecular structures, protein conformations, and continuous physical fields, which are inherently topological, geometric, or continuous. Several works have attempted to convert such data into text strings, such as using SMILES~\cite{smilesbert,chemberta2,chemformer} or SELFIES~\cite{krenn2020selfies} strings to represent molecular entities in molecular language models~\cite{molt5,llasmol}. This approach strikes a balance between linguistic reasoning and scientific insight to some extent. Nevertheless, linearizing molecular graphs into one-dimensional token sequences cannot explicitly expose the topological and geometric invariances central to chemical reasoning~\cite{threedmolm,unimol}, resulting in insufficiently refined molecular cognition and semantic noise that interferes with normal LLM reasoning.

\begin{figure}[ht]
    \centering
    \includegraphics[width=1\linewidth]{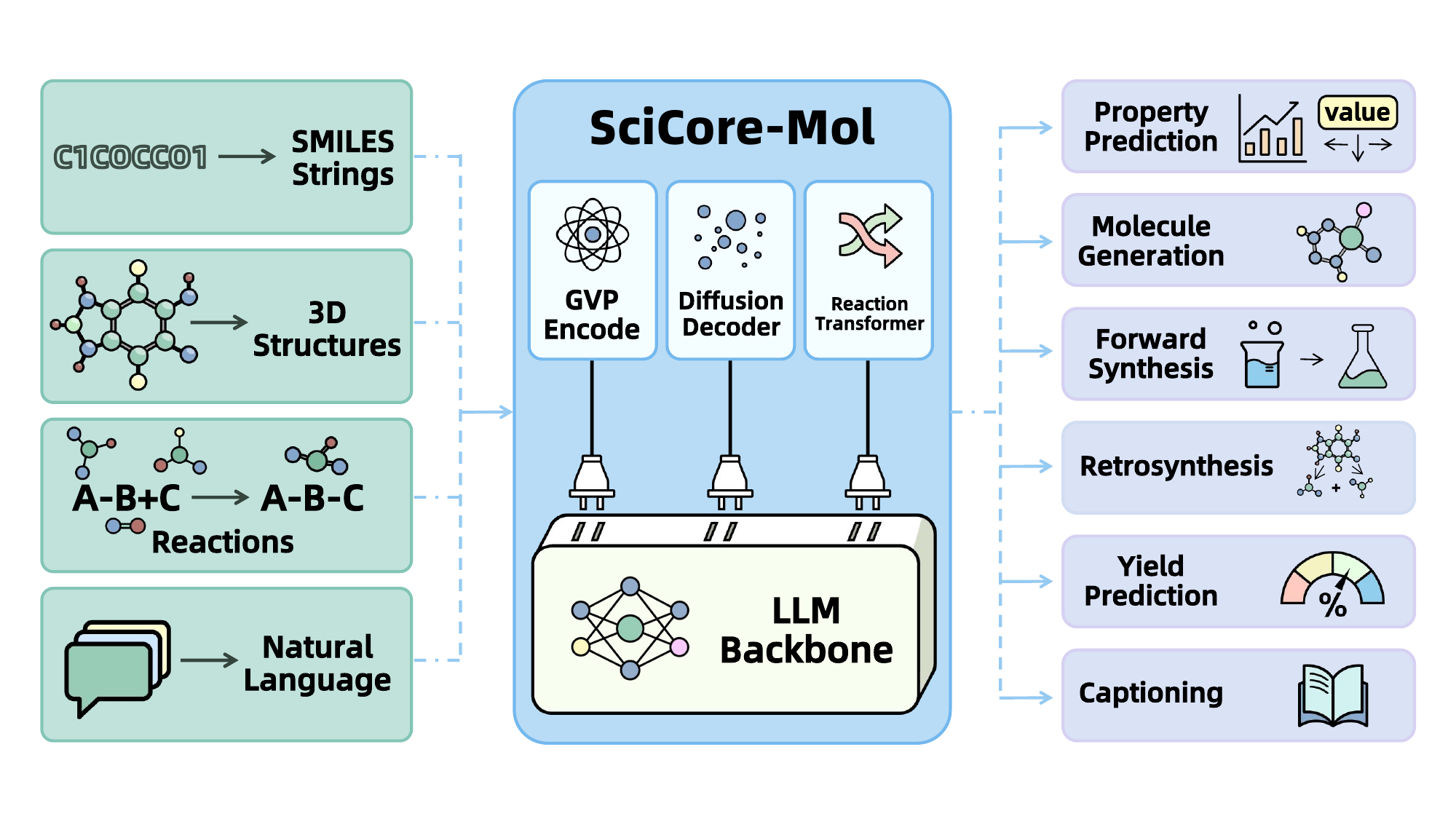}
    \caption{Overview of \ours{}. The GVP encoder, diffusion decoder, and reaction transformer correspond to the Topological Perception Module, Molecular Generation Module, and Reaction Sensing Module, respectively. \ours{} integrates these modules with an LLM backbone to support molecular property prediction, molecule generation, synthesis prediction, retrosynthesis, yield prediction, and captioning.}
    \label{fig1}
\end{figure}

Despite their strong general reasoning ability, LLMs remain unreliable as universal scientific solvers in structure-sensitive domains such as chemistry. Molecular properties depend on atomic connectivity, stereochemistry, conformational geometry, and inter-molecular interactions. A text-only LLM must infer these structural features indirectly from token co-occurrence, making its predictions vulnerable to spurious correlations. Recent chemistry benchmarks show that even strong frontier LLMs can fail on basic chemical reasoning tasks and produce overconfident predictions~\cite{mirza2025chembench}. These limitations motivate augmenting scientific LLMs with modules that explicitly encode molecular topology, geometry, and reaction-level numerical structure, rather than relying on text-based molecular representations alone.

A parallel line of work augments LLMs with specialized scientific tools including molecular graph encoders, diffusion generators, reaction predictors, and laboratory automation platforms~\cite{chemcrow,coscientist,liu2023moleculestm,ldmol}. These systems have substantially improved LLM usability in chemistry. However, in most existing systems the LLM functions primarily as a language interface while core scientific computation is carried out by external task-specific modules. Intermediate molecular, geometric, or numerical information must be compressed back into textual descriptions before being used by the LLM. This text-level interface introduces information loss and reasoning bottlenecks, especially for tasks that require structure-grounded perception, continuous molecular generation, or numerically sensitive reaction reasoning.

This motivates a tighter integration paradigm for scientific LLMs. Instead of treating scientific models as external black-box tools, we argue that heterogeneous scientific modules should become pluggable cognitive components of the LLM itself. In such a framework, the LLM remains responsible for natural-language interaction, task understanding, reasoning, and module coordination, while topology-aware, generation-aware, and reaction-aware modules provide specialized capabilities through learned hidden-state interfaces. This design preserves the flexibility and generality of LLM-centered reasoning while eliminating the information bottleneck of text-only tool feedback, and allows different scientific capabilities to be selectively activated and jointly optimized within a unified system.

We propose \ours{}, which consists of a LLM backbone and three pluggable external modules: a Topological Perception Module built on a GVP (Geometric Vector Perceptron) network~\cite{gvp} for spatial structure encoding, a Molecular Generation Module based on Diffusion Transformers (DiT)~\cite{dit}, and a Reaction Sensing Module built on a numerical-sensitive Transformer for inter-molecular reasoning. The LLM is responsible for language interaction, reasoning, and task coordination, determining when to invoke which module to enhance itself. The pluggable modules are integrated with the LLM backbone at the hidden embedding level, jointly performing information perception and decision-making, thereby alleviating the fundamental contradiction between linguistic reasoning and continuous scientific manifolds. To achieve effective module integration, we introduce a progressive training strategy: we first pre-train each module independently, then jointly align and optimize all modules together with the LLM on a high-quality domain corpus, and optionally fine-tune the integrated system for specific downstream tasks.

We evaluate \ours{} on molecular property prediction, conformation generation, chemical reaction product and yield prediction, and general chemistry knowledge benchmarks. Results show that \ours{} achieves superior performance on most tasks compared to baselines of similar scale, with an 8B-parameter system competitive with and in several dimensions surpassing much larger closed-source models. A recently released drug optimization dataset---collected after our pre-training cutoff---further confirms strong out-of-distribution generalization. We believe this framework can be extended to other scientific fields requiring heterogeneous data integration, advancing the development of scientific foundation models.

\section{Results}\label{sec:results}

\begin{figure*}[!t]
\centering
\includegraphics[width=\textwidth]{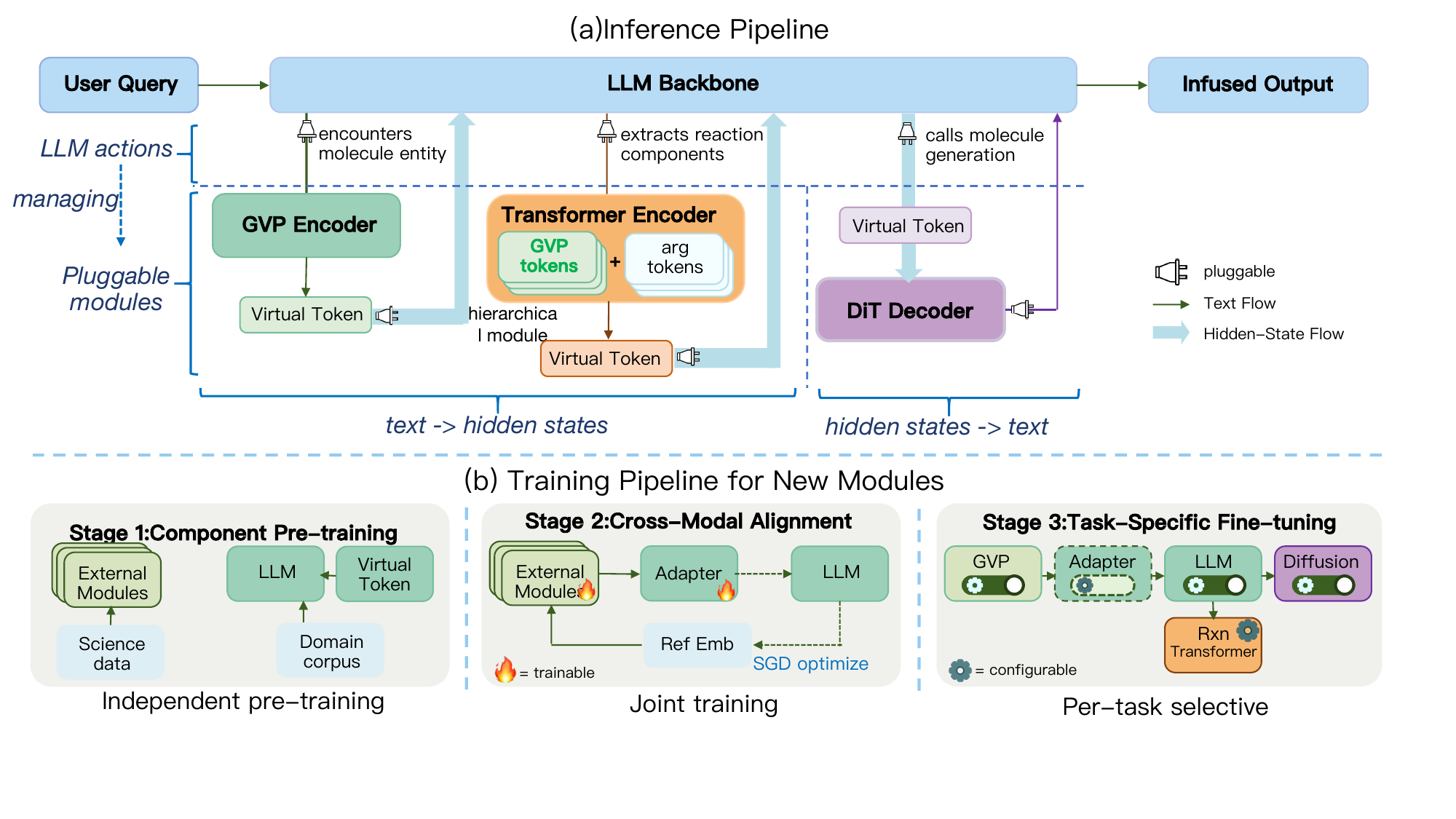}
\caption{\textbf{(a)} Inference pipeline of \ours{}. The GVP encoder, Reaction Transformer, and DiT decoder implement the Topological Perception Module, Reaction Sensing Module, and Molecular Generation Module, respectively. These pluggable modules exchange information with the LLM backbone through hidden-state interfaces. \textbf{(b)} Progressive training pipeline, including independent component pre-training, cross-modal alignment, and task-specific fine-tuning.}
\label{fig:architecture}
\end{figure*}

%--------------------------------------
\subsection{Overview of \ours{}}\label{sec:results_overview}
%--------------------------------------

We propose \ours{}, a molecular language model enhanced by three pluggable modules (Fig.~\ref{fig:architecture}):

\begin{enumerate}
  % \item \textbf{Topological Perception Module} that encodes 3D molecular geometry via a GVP network and injects the resulting embedding into the LLM's hidden-state sequence as a Virtual Structural Token.
   \item \textbf{Topological Perception Module} which uses a rotation-aware GVP network to encode 3D molecular geometry and injects the resulting structural embedding into the LLM hidden-state sequence as a
  Virtual Structural Token.
  \item \textbf{Molecular Generation Module} that generates molecules via latent diffusion conditioned on LLM hidden states, where the iterative denoising process (implemented by a DiT) is better suited than autoregressive token decoding for modeling globally coherent molecular structures.
  %\item \textbf{Molecular Generation Module} that generates molecules through latent diffusion conditioned on LLM hidden states.
  \item \textbf{Reaction Sensing Module}, a Transformer-based module that processes molecular reaction data hierarchically using GVP-derived molecular embeddings for reaction prediction and yield estimation.
\end{enumerate}

During inference, the LLM backbone processes all inputs as natural language and selectively activates external modules based on task context: when a molecular entity is detected in the input, the Topological Perception Module encodes its 3D conformation and injects the resulting embedding into the hidden-state sequence on the fly, enabling downstream tokens to attend to geometric information through the causal attention mask\textbf{;} when a generation task is \textbf{identified}, the Molecular Generation Module takes the LLM's hidden states as conditioning and produces molecular structures through iterative denoising in a continuous latent space\textbf{;} when a reaction-level query is encountered, the Reaction Sensing Module encodes multi-channel molecular tokens---combining structural, role, and stoichiometric information---and returns predicted products, yields, or missing reactants as additional context for continued reasoning. Module dispatch is controlled by the LLM itself through predefined special tokens that indicate whether topological perception, molecular generation, or reaction reasoning should be invoked at the current decoding step. All three modules share the same LLM backbone and can be independently enabled or disabled, making the system composable across task types. Architecture details and the progressive training strategy are described in Methods.

To comprehensively evaluate \ours{}, we organize results along five capability dimensions: \emph{Knowledge Core} (general chemistry knowledge), \emph{Mol-Text Translation} (bidirectional conversion between molecular representations and natural language), \emph{Molecule Generation} (\emph{de novo} molecular design and optimization), \emph{Quantitative Prediction} (property regression and yield estimation), and \emph{Synthesis Reasoning} (forward synthesis and retrosynthesis prediction). We first present the main comparison via per-model capability radar charts, followed by module-level ablation studies that isolate the contribution of each pluggable component.

%--------------------------------------
\subsection{Baseline Models}\label{sec:results_baselines}
%--------------------------------------

We compare \ours{} (an 8B LLM backbone and approximately 8.6B total parameters including all pluggable modules) with two groups of baselines: (1) strong closed-source models that represent high-performing general-purpose systems, and (2) open-source models at a comparable parameter scale.

\textbf{GPT-4o \& GPT-5}~\cite{hurst2024gpt, singh2025openai}. Closed-source large language models used as strong general-purpose baselines. Both models process molecular inputs as SMILES text without dedicated structural encoders.

\textbf{Intern-S1-mini}~\cite{intern-s1}\footnote{\url{https://huggingface.co/internlm/Intern-S1-mini}}. A science-specialized language model with strong reasoning capabilities. While not specifically designed for chemistry, it benefits from broad scientific pre-training and chain-of-thought reasoning.

\textbf{LlaSMol-Mistral-7B}~\cite{llasmol}. A chemistry-tuned model fine-tuned on the SMolInstruct dataset. It represents a text-centric paradigm for molecular tasks, processing molecular information as SMILES strings without geometric or reaction-specific modules.

\textbf{Qwen3-8B}~\cite{qwen3}\footnote{Qwen3-8B-Instruct: \url{https://huggingface.co/Qwen/Qwen3-8B}}. The pre-trained foundation model without additional chemistry-domain adaptation, serving as a general-purpose LLM baseline for chemical tasks.

\textbf{Qwen3-8B-Chem}. The Qwen3-8B backbone after continual pre-training on a chemistry-enriched corpus and supervised fine-tuning on molecular instructions, but without pluggable modules. This ablation separates the contribution of domain adaptation from that of the proposed modules.

%--------------------------------------
\subsection{Datasets and Metrics}\label{sec:results_datasets}
%--------------------------------------

\begin{figure*}[t]
\centering
\includegraphics[width=\textwidth]{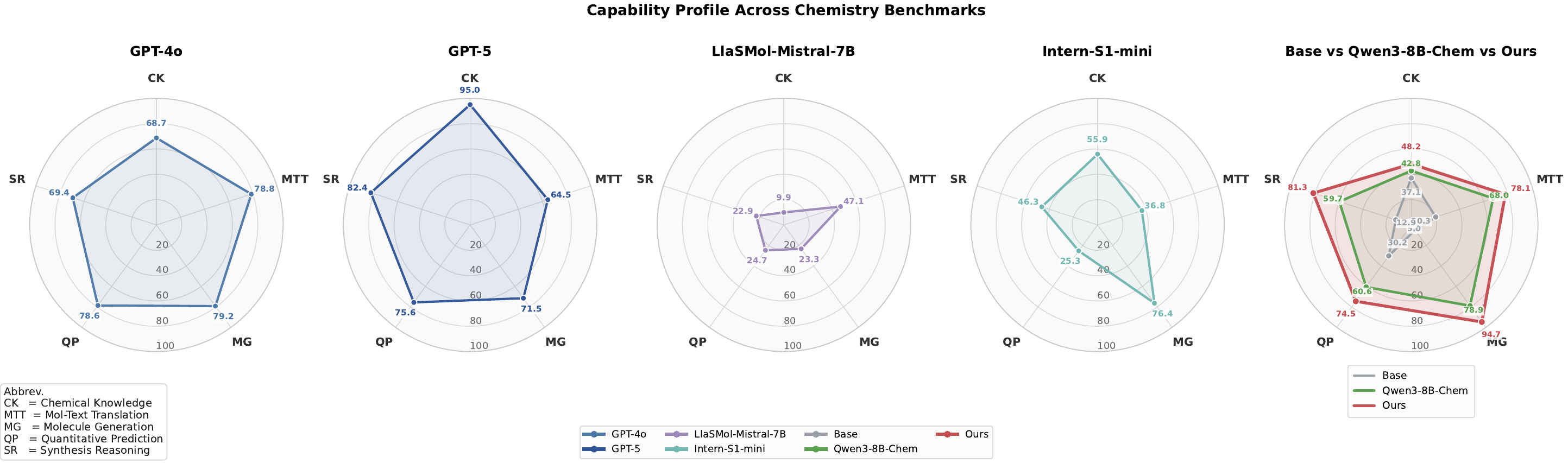}
\caption{Per-model capability radar charts across five evaluation dimensions. Raw benchmark metrics are normalized to $[0,100]$ via min--max scaling (Eq.~\ref{eq:normalization}); the normalization procedure and metric groupings are described in the \nameref{sec:app_evaluation} section. \ours{} achieves the most balanced and competitive profile overall.}\label{fig:radar}
\end{figure*}

We evaluate \ours{} on six benchmark suites covering complementary chemical tasks:

  \begin{itemize}
      \item \textbf{SMolInstruct}~\cite{llasmol}: a generative benchmark spanning six molecular task families, including name conversion, property prediction, molecule generation, forward synthesis, retrosynthesis, and captioning. Unlike multiple-choice benchmarks, these tasks require models to generate free-form outputs.

      \item \textbf{MMLU-Chemistry}~\cite{mmlu}: a subset of five chemistry-related subjects from the Massive Multitask Language Understanding benchmark, including high-school chemistry, college chemistry, organic chemistry, physical chemistry, and general science. We use this suite to examine whether domain adaptation preserves broad scientific reasoning ability.

      \item \textbf{ChemBench4K}~\cite{mirza2025chembench}: a multiple-choice chemistry benchmark covering product prediction, retrosynthesis, yield prediction, caption-to-molecule, and molecule-to-caption tasks. It provides a discriminative complement to the generative evaluation on SMolInstruct.

      \item \textbf{ORD}~\cite{ord}: a reaction-centered benchmark built from the Open Reaction Database, covering product prediction, joint product-and-yield prediction, standalone yield prediction, and reactant prediction. We use it to evaluate inter-molecular reasoning under realistic reaction conditions.

      \item \textbf{MoleculeNet}~\cite{moleculenet}: a standard benchmark for molecular property prediction, including classification tasks such as BBBP, Tox21, ClinTox, HIV, BACE, and SIDER, as well as regression tasks such as ESOL, FreeSolv, Lipo, and QM9. We use it primarily to ablate the Topological Perception Module.

      \item \textbf{DrugR}~\cite{drugr}: a drug optimization benchmark released after our pre-training data cutoff, which we use to evaluate out-of-distribution behavior of the Molecular Generation Module.
  \end{itemize}

Unless otherwise noted, metrics are reported according to task type. For \emph{molecular generation} tasks, including de novo design and SMILES-oriented generation in SMolInstruct, we report RDK fingerprint Tanimoto similarity and molecular validity~\cite{landrum2013rdkit}. For \emph{regression} targets, including physicochemical properties and continuous reaction-yield error, we report RMSE or MAE as specified by each dataset. For \emph{classification} targets, we report accuracy or F1. For \emph{text-generation} tasks such as molecule captioning, we report METEOR~\cite{banerjee2005meteor}. For \emph{reaction-outcome} evaluation, where models rank predicted yields against experimental outcomes, we report nDCG~\cite{jarvelin2002cumulated}. The prompt templates used for all benchmark datasets are provided in Section~\nameref{sec:app_prompts}.

%--------------------------------------
\subsection{Main Results}\label{sec:results_main}
%--------------------------------------

Figure~\ref{fig:radar} presents per-model capability radar charts across the five evaluation dimensions. We aggregate scores from multiple benchmarks, with full per-benchmark results reported in the \nameref{sec:app_evaluation} section. The capability scores are normalized to $[0,100]$ using the min--max scaling procedure in Eq.~\ref{eq:normalization}.

Several observations emerge. First, \ours{} achieves the most balanced capability profile among all evaluated models. Compared with open-source baselines at a similar parameter scale, including LlaSMol-Mistral-7B, Qwen3-8B, and Qwen3-8B-Chem, \ours{} consistently obtains higher scores across molecule generation, quantitative prediction, and synthesis reasoning, while maintaining competitive performance on the Knowledge Core dimension. This indicates that the gains are not merely due to chemistry-domain tuning of the LLM backbone, but arise from the complementary capabilities introduced by the pluggable structural, generative, and reaction-aware modules.

Second, although closed-source models such as GPT-4o and GPT-5 benefit from substantially larger model capacity and broader pre-training corpora, with their strongest advantages appearing in general chemistry knowledge and broad factual recall. In contrast, \ours{} approaches or surpasses them on application-oriented chemical tasks that require molecular structure perception, reaction-level reasoning, or controllable molecule generation. These results suggest that targeted modular training can provide useful inductive biases for chemistry-specific reasoning and generation, even without relying solely on model scale.

%--------------------------------------
\subsection{Topological Perception Module}\label{sec:results_gvp}
%--------------------------------------

Table~\ref{tab:sup-gvp} presents an ablation study evaluating the contribution of the GVP encoder on MoleculeNet property prediction tasks. We compare four model variants: Qwen3-8B base, \textbf{Ours w/o SFT, w/o GVP} (continual pre-training only), \textbf{Ours w/o GVP} (domain-adapted without the topological module), and the full \textbf{Ours} system.

As shown in Table~\ref{tab:sup-gvp}, the full \ours{} system consistently outperforms the variants without GVP, with the largest gains on tasks where 3D information provides decisive signal (e.g., Tox21, HIV classification) and on regression targets that depend on molecular geometry (e.g., QM9). These findings support two conclusions: (1)~the GVP encoder injects complementary geometric information that is difficult to recover from SMILES alone, and (2)~the projection adapter effectively aligns geometric representations with the LLM hidden space.

\begin{table*}[t]
\centering
\small
\setlength{\tabcolsep}{5pt}
\caption{GVP ablation on MoleculeNet property prediction tasks. F1 ($\uparrow$, higher is better); MAE ($\downarrow$, lower is better). Best result in each column is \textbf{bolded}.}
\label{tab:sup-gvp}
\begin{tabular}{lccccc|cccc}
\toprule
\textbf{Model} &
\multicolumn{5}{c|}{\textbf{MoleculeNet Classification (F1)$\uparrow$}} &
\multicolumn{4}{c}{\textbf{MoleculeNet Regression (MAE)$\downarrow$}} \\
\cmidrule(lr){2-6}\cmidrule(lr){7-10}
& \textbf{BBBP} & \textbf{Tox21} & \textbf{ClinTox} & \textbf{HIV} & \textbf{BACE}
& \textbf{ESOL} & \textbf{FreeSolv} & \textbf{Lipo} & \textbf{QM9} \\
\midrule
Qwen3-8B                           & 0.47          & 0.27          & 0.50          & 0.25          & 0.61          & 1.11          & 0.72          & 1.00          & 0.92 \\
Ours w/o SFT, w/o GVP              & \textbf{0.53} & 0.35          & 0.39          & 0.56          & 0.63          & 0.98          & \textbf{0.71} & 0.92          & 0.86 \\
Ours w/o GVP                       & 0.51          & 0.53          & 0.50          & 0.70          & 0.60          & 0.93          & 0.97          & 0.89          & 0.90 \\
\textbf{Ours}                      & 0.49          & \textbf{0.85} & \textbf{0.53} & \textbf{0.94} & \textbf{0.65} & \textbf{0.88} & 0.81          & \textbf{0.86} & \textbf{0.78} \\
\bottomrule
\end{tabular}
\end{table*}

%--------------------------------------
\subsection{Molecular Generation Module Ablation}\label{sec:results_diffusion}
%--------------------------------------

Table~\ref{tab:avg-main-reward} compares molecular generation quality across different model configurations on a drug optimization task, reporting average main reward (higher indicates better pharmaceutical properties) and structural similarity (RDK-FTS).

As shown in Table~\ref{tab:avg-main-reward}, \ours{} improves average main reward over the Qwen3-8B base and the domain-adapted Qwen3-8B-Chem, suggesting that the diffusion-based generation module provides gains beyond language-only adaptation. The RDK-FTS similarity metric also shows a trade-off between reward-oriented optimization and scaffold preservation, suggesting that editing-oriented settings may benefit from stronger structure-preserving guidance schedules.

\begin{table}[t]
\centering
\small
\caption{Comparison of average main reward and structure similarity (RDK-FTS) across different model variants. Higher is better.}
\label{tab:avg-main-reward}
\setlength{\tabcolsep}{5pt}
\begin{tabular}{lcc}
\toprule
\textbf{Model} & \textbf{Avg. Main Reward} & \textbf{RDK-FTS} \\
\midrule
GPT-4o                 & 0.2545 & 0.7276 \\
GPT-5                  & 0.2508 & 0.7436 \\
Intern-S1-mini         & 0.2097 & 0.7263 \\
Intern-S1-mini (SFT)   & 0.2085 & 0.8625 \\
Qwen3-8B               & 0.1465 & 0.8795 \\
Qwen3-8B-Chem          & 0.2202 & 0.3823 \\
\ours{}                & 0.2380 & 0.5442 \\
\bottomrule
\end{tabular}
\end{table}

%--------------------------------------
\subsection{Result Analysis}\label{sec:results_analysis}
%--------------------------------------

From the experimental results, we draw three main findings:

\begin{enumerate}
    \item \textbf{Modular augmentation preserves general capabilities.} Integrating three specialized modules does not degrade Knowledge Core scores or general reasoning ability. The combination of KL-regularized continual pre-training and progressive training effectively prevents catastrophic forgetting, allowing the model to gain molecular expertise as an additive enhancement rather than a trade-off.

    \item \textbf{Geometric features provide decisive advantages on structure-sensitive targets.} The GVP ablation (Table~\ref{tab:sup-gvp}) shows that adding 3D geometric embeddings consistently improves performance on multiple MoleculeNet tasks, supporting the view that pure text-based LLMs cannot reliably capture the spatial information required by many physicochemical prediction problems.

    \item \textbf{Reaction-level reasoning requires dedicated architectural support.}
    The ORD ablation in Table~\ref{tab:ord-main} shows that removing the Reaction Sensing Module causes severe degradation in product prediction validity, joint product-and-yield validity, and reactant prediction validity. This confirms that reaction-level reasoning cannot be reliably recovered from text-only reaction descriptions alone, and instead benefits from structured modeling over molecular roles, stoichiometric quantities, and inter-molecular interactions.
\end{enumerate}

The domain-adapted LLM without pluggable modules (Qwen3-8B-Chem) already improves over the base model on several tasks, suggesting that continual pre-training and instruction tuning on chemical corpora provide useful domain knowledge. However, the remaining gap between Qwen3-8B-Chem and the full \ours{} system suggests that pluggable modules add capabilities that are difficult to obtain through text-based training alone.

\begin{table*}[t]
\centering
\scriptsize
\setlength{\tabcolsep}{3pt}
\renewcommand{\arraystretch}{1.25}
\newcommand{\molpanel}[2]{%
\begin{minipage}{0.96\linewidth}
\centering
\includegraphics[height=1.55cm,keepaspectratio]{#1}\\[-2pt]
\scriptsize #2
\end{minipage}}
\begin{tabular}{p{0.12\textwidth} p{0.28\textwidth} p{0.28\textwidth} p{0.28\textwidth}}
\toprule
 &
\textbf{(a) Molecular Generation} &
\textbf{(b) Product + Yield Prediction} &
\textbf{(c) Molecular Captioning} \\
\midrule
\textbf{Input} &
\textbf{\textcolor{blue}{Q:}} Generate a tankyrase-inhibitor-like thiopyranopyrimidine with a para-\texttt{CF3} phenyl substituent. &
\textbf{\textcolor{blue}{Q:}} Given two reactants, predict the coupled product and yield. &
\textbf{\textcolor{blue}{Q:}} Provide a brief overview of this molecule. \newline
\molpanel{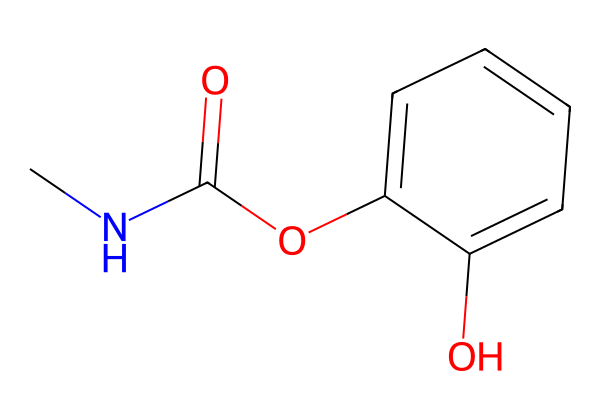}{Input molecule: aromatic carbamate.} \\
\midrule
\textbf{Ground Truth} &
\molpanel{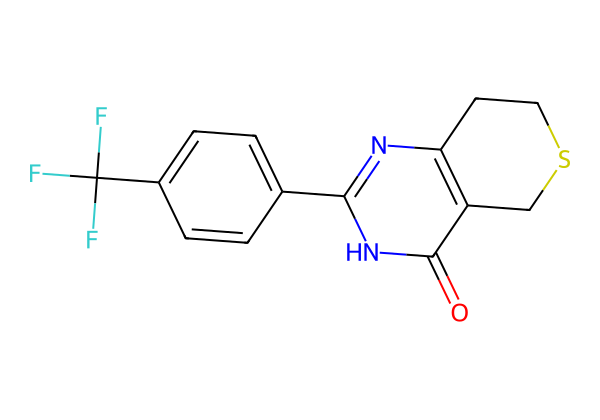}{Reference scaffold: thiopyranopyrimidine + para-\texttt{CF3} phenyl.} &
\molpanel{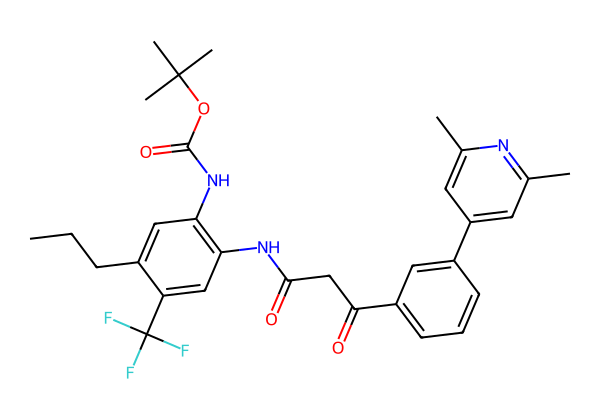}{Reference coupled product; yield \textbf{76.00\%}.} &
\textit{``Carbamate pesticide derived from carbamic acid, used in agricultural or residential pest control.''} \\
\midrule
\textbf{Baseline} &
\textbf{GPT-5:} \textcolor{red}{sim.\ 0.537 (scaffold drift).} \newline
\textbf{Intern-S1-mini:} \textcolor{red}{sim.\ 0.043 (trivial analogue).} &
\textbf{GPT-4o/GPT-5:} \textcolor{red}{\textit{``\ldots reactants remain unreacted\ldots''} (incomplete product; no valid yield).} \newline
\textbf{Qwen3-8B-Chem:} \textcolor{red}{aryl ketone-like product; yield est.\ $>$90\% (overconfident).} &
\textbf{GPT-4o/Qwen3-8B:} \textcolor{red}{\textit{``\ldots typical aromatic ester used as plasticizer\ldots''} (wrong functional group).} \\
\midrule
\textbf{\ours{}} &
\molpanel{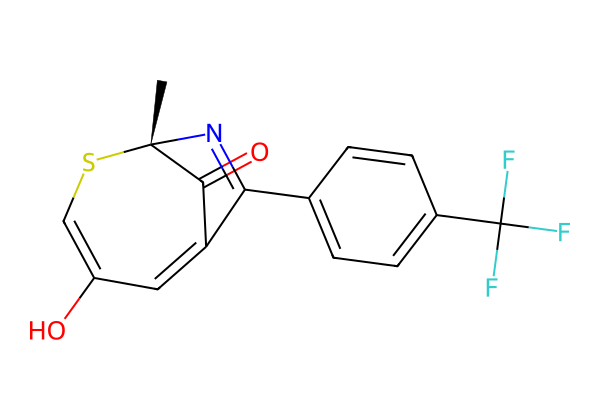}{\textbf{Ours:} \textcolor{green!45!black}{sim.\ 0.607; core scaffold preserved.}} &
\textbf{Ours:} \textcolor{green!45!black}{Correct coupled product recovered; predicted yield \textbf{57.60\%} (ranking score 0.819).} &
\textbf{Ours:} \textcolor{green!45!black}{\textit{``\ldots carbamate motif\ldots derived from carbamic acid\ldots commonly applied in pest control.''}} \\
\bottomrule
\end{tabular}
\caption{Representative case studies across three inference pathways. Molecule renderings replace SMILES strings. Quoted text shows key excerpts from actual model outputs; analysis is in the main text.}
\label{tab:case_studies}
\end{table*}

%--------------------------------------
\subsection{Case Study}\label{sec:results_case_study}
%--------------------------------------

\ours{} supports three qualitatively different inference pathways, illustrated in Table~\ref{tab:case_studies}.

\textbf{(a) Molecular generation.} Given the prompt ``generate a tankyrase-inhibitor-like thiopyranopyrimidine with a para-\texttt{CF3} phenyl substituent'', baseline LLMs either drift far from the reference scaffold or collapse to trivial analogues (GPT-5: similarity 0.537; Intern-S1-mini: 0.043). \ours{} leverages the diffusion-based Molecular Generation Module to produce a candidate that better preserves the thiopyranopyrimidine core and substituent pattern (similarity 0.607), illustrating that the module can follow fine-grained structural instructions.

\textbf{(b) Product and yield prediction.} For a coupling reaction with experimental yield 76.00\%, general-purpose LLMs fail to output a single coherent product---GPT-4o and GPT-5 return separated reactants or incomplete product sets, while Qwen3-8B-Chem produces an unrelated aryl ketone-like structure with a yield estimate far from the target. \ours{} recovers the correct coupled-product pattern and predicts 57.60\% yield (ranking score 0.819), confirming that the Reaction Sensing Module improves both structural validity and quantitative accuracy.

\textbf{(c) Molecular captioning.} When asked to ``provide a brief overview'' of an aromatic carbamate, baseline models (GPT-4o, Qwen3-8B) describe it as a generic aromatic ester or plasticizer. \ours{}, informed by the GVP-encoded 3D structure, correctly identifies the carbamate motif, notes its carbamic acid derivation, and links it to pesticide applications.

Together, these cases demonstrate that the three pluggable modules yield more faithful and chemically grounded behavior in realistic open-ended scenarios.

\section{Discussion}\label{sec:discussion}

In this work, we present \ours{}, a modular molecular language model that augments an LLM backbone with three pluggable modules: a Topological Perception Module for 3D geometric encoding, a Molecular Generation Module based on latent diffusion, and a Reaction Sensing Module for inter-molecular reasoning. Each module is integrated at the hidden-state level through standardized interfaces. The progressive training strategy aligns these heterogeneous modules with the shared language backbone while preserving general reasoning capabilities.

Our experiments show that a unified architecture can support molecular understanding, generation, and reaction reasoning within a single system. The radar chart comparison indicates that \ours{} exhibits the most balanced capability profile among models in similar scale. It even surpasses strong closed-source models in some dimensions while allowing further tuning. This observation suggests that deep, embedding-level integration of specialized modules is more effective than relying solely on text-based molecular representations. 

Particularly, the model's out-of-distribution generalization on the DrugR benchmark is an encouraging result, because this is a newly released dataset that eliminates the possibility of data leakage during the pre-training phase, and evaluates the model's ability to flexibly optimize drug properties for new molecule types and scaffolds. This suggests that our model acquires transferable structural priors from the LLM's chemical knowledge rather than depending only on memorization, highlighting its potential for practical drug design scenarios. Additional evidence comes from our few-shot prompting experiments. When provided with a small number of in-context examples, \ours{} can make effective use of new demonstrations and adapt its reasoning behavior accordingly. In contrast, several baseline models do not benefit as consistently from the same setting, suggesting that the proposed architecture is better suited for compositional reasoning over molecular structures rather than surface-level pattern matching or knowledge memory.

Beyond predictive performance, the modular design also offers practical advantages. Each module can be upgraded independently. For example, the GVP encoder in the Topological Perception Module can be replaced with a stronger 3D architecture without retraining the entire system. In deployment settings that require only a subset of capabilities, unused modules can also be detached to reduce computational overhead. Further, the hierachical architecture of external modules can strengthen the multi-molecular processing and numerical capability, revealing the potential of supporting complicated systems such as polymers and crystal structures.

Despite these strengths, several limitations remain. The Molecular Generation Module still has room for improvement in structural fidelity. The current system is designed for small molecules and does not yet support larger entities such as proteins. Numerical reasoning for yield and stoichiometric calculations, although strengthened by the Reaction Sensing Module, remains challenging in precision-critical settings. In addition, the three-stage training pipeline introduces extra complexity and requires careful coordination across stages. Future work may extend the module library to protein and polymer modalities, incorporate reinforcement learning from chemical feedback such as docking scores or synthetic accessibility, scale the framework to larger LLM backbones, and integrate real-time experimental feedback for closed-loop molecular design.

\section{Conclusion}\label{sec:conclusion}

We demonstrate that augmenting LLMs with specialized and deeply integrated molecular modules yields a unified system capable of strong performance across diverse chemical tasks. The plug-and-play architecture supports flexible deployment and independent module evolution, while the progressive training strategy enables stable optimization across heterogeneous objectives. We believe this framework offers a general template for equipping LLMs with scientific expertise through decoupled, pluggable modules, and it can be naturally extended to other scientific domains that require heterogeneous data integration.

\section*{Code and Data Availability}

The code and training scripts for \ours{} are publicly available at \url{https://github.com/OpenBMB/SciCore-Mol}. Model weights are hosted on Hugging Face at \url{https://huggingface.co/openbmb/SciCore-Mol}. An interactive demo can be accessed at \url{https://chenyxyx-scicore-mol.hf.space/}. Benchmark datasets used in this work (SMolInstruct, MoleculeNet, ChemBench4K, ORD, MMLU-Chemistry, DrugR) are available from their respective original sources cited in the Methods section.
% sections/4_methods.tex
\section{Methods}\label{sec:method}

%--------------------------------------
\subsection{Related Work}\label{sec:related_work}
%--------------------------------------

\subsubsection{Molecular Representation Learning}

Molecular representation learning has evolved from sequence-based models to graph neural networks and, more recently, geometry-aware 3D architectures. Early sequence-based methods treat molecules as linearized strings and learn transferable representations through large-scale pre-training, as exemplified by SMILES-BERT, ChemBERTa, and Chemformer \cite{smilesbert,chemberta2,chemformer}. While such approaches interface naturally with Transformer-based language modeling pipelines, they inevitably compress molecular topology and geometry into one-dimensional token sequences. To better preserve structural information, later work introduced graph-based models that explicitly encode molecular connectivity, such as GCN, GIN, and Graphormer \cite{gcn,gin,graphormer}, followed by 3D architectures that further incorporate atomic coordinates, directional relations, and geometric constraints, including GVP, SchNet, EGNN, PaiNN, DimeNet, and Uni-Mol \cite{gvp,schnet,egnn,painn,dimenet,unimol}. These geometry-aware models have shown clear advantages on structure-sensitive tasks. In addition, contrastive molecular pre-training methods such as MolCLR further improve representation generalization by leveraging graph augmentations \cite{molclr}.

Despite their strong inductive biases, most existing molecular encoders are developed as standalone backbones or task-specific predictors. As a result, they do not naturally provide the flexible reasoning and compositional generalization offered by large language models. This gap motivates approaches that preserve molecular topology and geometry while coupling them more tightly with an LLM-centered reasoning system.

\subsubsection{Generative Models for Molecules}

Molecular generation has also undergone a substantial transition, moving from variational autoencoders, generative adversarial networks, and flow-based models toward diffusion-based paradigms \cite{jtvae,moflow,ddpm}. Earlier methods established the feasibility of learning molecular distributions, but often faced limitations in generation stability, controllability, or the faithful modeling of chemical and geometric constraints. In contrast, diffusion models have recently emerged as a powerful alternative due to their stable optimization behavior and strong generation quality.

Recent molecular diffusion models have demonstrated promising results in both discrete graph generation and continuous 3D structure generation. Representative examples include GeoDiff, equivariant diffusion models for molecular graphs, DiGress, MiDi, GeoLDM, and LDMol \cite{geodiff,edm,digress,midi,geoldm,ldmol}. These methods are particularly attractive for chemistry because they provide a natural framework for modeling structured uncertainty while maintaining chemical validity and geometric consistency. However, most prior work treats molecular generation as an isolated objective. Even when language conditioning is introduced, the interaction between the generator and the language model is often limited to shallow prompt-level control. In realistic scientific workflows, generation must be coordinated with molecular understanding, reaction constraints, and quantitative reasoning, which calls for a more integrated design.

\subsubsection{Large Language Models in Chemistry}

Large language models have recently been adapted to chemistry through text-centric, multimodal, and tool-augmented paradigms. Text-centric approaches cast molecular tasks into unified sequence generation problems, enabling question answering, molecule captioning, name conversion, property prediction, and reaction prediction within a common language modeling framework. Representative models and resources include MolT5, ChemLLM, Galactica, BioT5, Mol-Instructions, SMolInstruct, and LLaSMol \cite{molt5,chemllm,galactica,biot5,molinstructions,llasmol}. These works demonstrate that LLMs can provide a flexible natural-language interface for chemistry and can unify multiple molecular tasks under an instruction-following format.

Nevertheless, most text-centric chemistry LLMs still rely heavily on textual molecular abstractions such as SMILES or SELFIES. As a result, molecular topology, stereochemistry, and geometry must be recovered implicitly from token sequences. This is inefficient for structure-sensitive tasks and can lead to unreliable reasoning when the target property depends on spatial conformation, reaction context, or numerical quantities. This limitation helps explain why domain-adapted LLMs can improve over general-purpose LLMs but still struggle to fully replace specialized molecular encoders, reaction models, or generative models in many scientific settings.

Multimodal chemistry LLMs attempt to address this gap by aligning language representations with molecular graphs, 3D structures, or biochemical knowledge. Representative efforts include MoleculeSTM, MolCA, Git-Mol, MolFM, 3D-MoLM, and mCLM \cite{liu2023moleculestm,liu2024molca,gitmol,molfm,threedmolm,mclm}. These methods show that incorporating molecular structures can improve molecule-text retrieval, captioning, open-ended molecular question answering, and text-guided editing. However, most of them focus on pairwise cross-modal alignment, modality conversion, or a limited set of downstream tasks. They do not fully unify molecular perception, molecule generation, reaction computation, and language reasoning within a single internally coordinated architecture. This motivates architectures in which specialized molecular modules interact with the language backbone through learned internal interfaces rather than through text alone.

Closely related efforts in materials science, such as MatterChat, align crystal-structure embeddings from a pretrained interatomic-potential model with an LLM for inorganic material property prediction and material question answering \cite{matterchat}. In contrast, \ours{} targets molecular chemistry rather than crystalline materials and integrates multiple pluggable cognition modules---geometry-aware molecular perception, latent molecular generation, and reaction-level quantitative reasoning---within one hidden-state interface.

Although these efforts have significantly improved conversational chemistry and cross-modal modeling, existing chemistry LLMs remain largely dominated by textual abstractions or relatively shallow cross-modal alignment. They are effective on question answering and straightforward modality conversion, but remain limited in structure-grounded reasoning, numerically sensitive prediction, and unified handling of molecular understanding, generation, and reaction computation. This limitation suggests the need for architectures in which specialized molecular modules interact with the language backbone through learned internal interfaces rather than text alone.

\subsubsection{LLM Agents and Tool-Augmented Reasoning}

Another closely related direction augments large language models with external tools. Tool-augmented systems extend LLMs beyond pure next-token prediction by allowing them to invoke retrieval systems, calculators, search engines, domain-specific software, and even laboratory automation platforms. In chemistry, such systems have shown promise in synthesis planning, molecular analysis, literature interaction, and experimental decision support. Representative examples include ChemCrow, Coscientist, ChatMol, and ChemAgent \cite{chemcrow,coscientist,chatmol,chemagent}. These systems demonstrate the practical value of LLMs as workflow coordinators, especially when tasks can be decomposed into explicit tool calls, symbolic computation, retrieval, or multi-step planning.

However, most existing agent frameworks rely on loose coupling: the LLM invokes an external tool, receives textual feedback, and then continues reasoning in language space. This interface is often insufficient for chemistry, where the core objects are fundamentally topological, geometric, continuous, and numerical. Dense molecular conformations, reaction states, stoichiometric quantities, and intermediate embeddings are difficult to faithfully transmit through natural-language feedback alone. As a result, tool-level interfaces may introduce information loss and create reasoning bottlenecks, even when the external tools themselves are accurate.

SciCore-Mol differs from these tool-augmented systems by integrating specialized scientific modules into the LLM's internal representation space. The modules are not treated as external black boxes, but as pluggable cognitive components that can be selectively invoked during inference and jointly aligned with the LLM backbone during training. This hidden-state-level integration allows molecular perception, molecular generation, and reaction reasoning to participate directly in the construction of the model's internal context, thereby combining the flexibility of LLM-centered reasoning with the inductive biases of specialized scientific architectures.

Overall, prior work has advanced molecular representation learning, molecular generation, chemistry-oriented LLMs, and tool-augmented reasoning largely in parallel. What remains underexplored is a unified framework that combines these capabilities without collapsing molecular structure into text or treating scientific modules as purely external black boxes. Our work is motivated by this gap and explores a modular architecture in which molecular perception, generation, and reaction reasoning are integrated with an LLM through learned interfaces.

The LLM backbone is continually pre-trained on a chemistry-enriched corpus
totaling $\sim$400M tokens, mixing domain-specific and general-purpose text
at a 1:1 ratio. This balance is chosen to substantially expand the model's
chemical knowledge while preserving its general reasoning capabilities.

The chemistry domain portion ($\sim$200M tokens) consists of two sources:
\begin{itemize}
    \item \textbf{Chemical literature} ($\sim$52M tokens): Full-text chemistry
    papers covering organic chemistry, biochemistry, and materials science,
    providing structured scientific knowledge and SMILES-in-context descriptions.
    \item \textbf{Open Reaction Database (ORD)}~\cite{ord} ($\sim$148M tokens):
    A large-scale, machine-readable repository of chemical reaction records,
    including reactants, reagents, conditions, products, and experimental
    procedures, which directly supports the Reaction Transformer module's
    pre-training signal.
\end{itemize}

The general-purpose portion ($\sim$200M tokens) is drawn from the Nemotron
pre-training data collection~\cite{nemotron}, comprising four components:
\begin{itemize}
    \item \textbf{Nemotron-Pretraining-CC} ($\sim$50M tokens): Curated
    Common Crawl web text across three quality-filtered subsets.
    \item \textbf{Nemotron-Pretraining-CC-Math} ($\sim$50M tokens):
    Mathematical and quantitative reasoning content extracted from web corpora,
    supporting numerical inference capabilities.
    \item \textbf{Nemotron-Pretraining-Code-Synthetic} ($\sim$50M tokens):
    Synthetically generated code samples that reinforce structured reasoning
    and algorithmic thinking.
    \item \textbf{Nemotron-Pretraining-SFT} ($\sim$50M tokens): Instruction-
    formatted text from three subsets, maintaining the model's instruction-
    following ability during continual pre-training.
\end{itemize}

All text is deduplicated and quality-filtered prior to mixing. A
KL-divergence regularization term $\mathcal{L}_{\mathrm{KL}} =
D_{\mathrm{KL}}(p_\theta \| p_{\mathrm{ref}})$ against the frozen Qwen3
base model is applied throughout continual pre-training to prevent
catastrophic forgetting of general capabilities.

%--------------------------------------
\subsection{Fine-Tuning Data}\label{sec:sft_data}
%--------------------------------------
For supervised fine-tuning, we construct a multi-task instruction corpus of 300K instruction--response pairs ($\sim$245M tokens) at a 2.5:1 general-to-domain ratio. The general-purpose portion (214,285 samples, 71.4\%) is sampled from Nemotron-Post-Training-v2~\cite{nemotron}, a diverse instruction-following dataset covering general-domain conversation, reasoning, and scientific question answering. All samples are filtered to a maximum of 2,048 tokens per conversation.

The chemistry domain-specific portion (85,715 samples, 28.6\%) is sourced from SMolInstruct~\cite{smolinstruct} and covers five task categories:
\begin{itemize}
    \item \textbf{Reaction synthesis} (36,083 samples): Forward synthesis (18,144 samples) and retrosynthesis (17,939 samples), converting between reactants and products using reaction SMILES.
    \item \textbf{Name conversion} (38,995 samples): Bidirectional translation among SMILES strings, IUPAC names, and molecular formulas across four subtasks: SMILES-to-formula (10,094), IUPAC-to-formula (9,978), SMILES-to-IUPAC (9,630), and IUPAC-to-SMILES (9,293).
    \item \textbf{Molecule generation} (4,240 samples): Text-conditioned de novo molecular design, where outputs are routed to the diffusion decoder rather than decoded autoregressively as SMILES strings.
    \item \textbf{Molecule captioning} (4,162 samples): Generating natural-language descriptions of molecular structures, integrating GVP-encoded 3D representations with language generation.
    \item \textbf{Property prediction} (2,235 samples): Binary and regression-based physicochemical property tasks including lipophilicity, clinical toxicity, and aqueous solubility from MoleculeNet~\cite{moleculenet}.
\end{itemize}

Task balancing is applied using temperature-based allocation with $\alpha = 0.5$, with binary classification tasks further balanced by Yes/No label to mitigate class imbalance. Cross-file deduplication is performed using SHA-1 fingerprints of normalized message sequences. This corpus drives Stage~II joint multi-task training (the ~\nameref{sec:stage2} section), with the general portion maintaining instruction-following capability throughout.

%--------------------------------------
\subsection{Architecture}\label{sec:architecture}
%--------------------------------------

\subsubsection{Topological Perception Module}\label{sec:gvp_module}

SMILES strings encode molecular connectivity but lose 3D spatial information and relationships between molecules, and tokenizers are difficult to align directly for encoding, allowing LLM to understand the properties of the entire molecule. To provide geometric features for LLM, we employed a GVP network to encode the 3D conformation (generated via RDKit) as SE(3)-equivariant representation, integrating both scalar atom-level properties (atomic number, formal charge, aromaticity, ring membership, etc.) and 3D coordinate vectors through message passing, and align the resulting molecular representation with the hidden state of LLM.

\paragraph{GVP Encoder.}
Given a molecular graph $\mathcal{G} = (\mathcal{V}, \mathcal{E})$, each node carries scalar features $\mathbf{s}_v$ and vector features $\mathbf{V}_v \in \mathbb{R}^{\nu \times 3}$. A GVP layer transforms these as:
\begin{align}
    \mathbf{V}^{h} &= W_V \mathbf{V}, \\
    \mathbf{s}' &= \sigma_s\!\left(W_s\!\left[\mathbf{s}\,\Vert\,\|\mathbf{V}^{h}\|_2\right] + \mathbf{b}_s\right), \\
    \mathbf{V}' &= g(\mathbf{s}') \odot \mathbf{V}^{h}.
\end{align}
No bias is applied to vector channels, preserving SE(3)-equivariance. Four GVP message-passing layers with hidden dimensions $(256, 16)$ followed by global pooling produce a graph-level representation $\mathbf{h}_{\mathrm{geo}} \in \mathbb{R}^{256}$.

\paragraph{MLP Adapter.}
A two-layer MLP projects the geometric representation into the LLM hidden space ($d = 3{,}072$):
\begin{equation}
    \mathbf{h}^{\mathrm{mol}}_k = W_2\,\mathrm{ReLU}(W_1 \mathbf{h}_{\mathrm{geo}} + \mathbf{b}_1) + \mathbf{b}_2 \in \mathbb{R}^{d},
\end{equation}
with projection dimensions $256 \to 2{,}048 \to 3{,}072$. The projected $\mathbf{h}^{\mathrm{mol}}_k$ is appended to the hidden-state sequence as a \emph{Virtual Structural Token}:
\begin{equation}
    \tilde{\mathbf{H}}_{\mathbf{X}} = \mathrm{Concat}\!\left(\mathbf{H}_{\mathbf{X}},\, \mathbf{h}^{\mathrm{mol}}_1,\, \dots,\, \mathbf{h}^{\mathrm{mol}}_K\right),
\end{equation}
so that subsequent self-attention layers attend to 3D geometric information alongside textual tokens. A dedicated chemical entity detector scans the input, and when a molecular entity is detected, a \emph{virtual-step} is triggered on the fly---encoding and inserting the structural token without requiring all molecular features to be pre-computed before decoding.

\subsubsection{Geometric Optimization Module}\label{sec:diffusion_module}

Directly generating SMILES through the autoregressive decoding of an LLM often leads to low validity, because its tokenizer is not designed around SMILES and a single misplaced character can invalidate the entire output. To address this issue, we integrate a latent diffusion module that operates in a continuous molecular latent space, where a pre-trained SMILES autoencoder maps molecules to compact latent representations $\mathbf{z}_0 \in \mathbb{R}^{127 \times 64}$ and remains frozen throughout training.

\paragraph{Diffusion Process.}
Let $\alpha_t = 1-\beta_t$ and $\bar\alpha_t = \prod_{s=1}^{t}\alpha_s$. The full diffusion pipeline—forward noising, conditioned denoising, reference-guided bridge initialization, and classifier-free guidance (CFG)—is defined jointly:
\begin{gather}
    q(\mathbf{z}_t \mid \mathbf{z}_{t-1}) = \mathcal{N}\!\left(\sqrt{1-\beta_t}\,\mathbf{z}_{t-1},\,\beta_t\mathbf{I}\right), \label{eq:diff_fwd} \\[2pt]
    \hat{\boldsymbol\epsilon} = \epsilon_\theta(\mathbf{z}_t,\,t,\,\mathbf{c}), \qquad \mathbf{c} = \mathrm{TextProj}(\mathbf{H}_{\mathrm{LLM}}), \label{eq:diff_denoise} \\[2pt]
    \mathbf{z}_t^{\mathrm{bridge}} = \sqrt{\bar\alpha_t}\,\mathrm{Enc}(x_{\mathrm{src}}) + \sqrt{1-\bar\alpha_t}\,\boldsymbol\epsilon, \label{eq:diff_bridge} \\[2pt]
    \hat{\boldsymbol\epsilon}_{\mathrm{cfg}} = \epsilon_\theta(\mathbf{z}_t,t,\varnothing) + s\bigl[\epsilon_\theta(\mathbf{z}_t,t,\mathbf{c}) - \epsilon_\theta(\mathbf{z}_t,t,\varnothing)\bigr]. \label{eq:diff_cfg}
\end{gather}
Eq.~\eqref{eq:diff_fwd} gives the forward noising; Eq.~\eqref{eq:diff_denoise} is the DiT denoiser conditioned on LLM semantics; Eq.~\eqref{eq:diff_bridge} initializes from a source molecule for scaffold-preserving drug editing; Eq.~\eqref{eq:diff_cfg} applies CFG with guidance scale $s$. The final denoised latent is decoded to SMILES via the frozen autoencoder.

\subsubsection{Molecular Interaction Module}\label{sec:interaction_module}

Although \ours{} handles single-molecule tasks reasonably well with the topological perception module, it is less effective at modeling inter-molecular reactions,  particularly for numerical tasks such as yield prediction where precise quantitative reasoning is required. However, LLMs can be trained to recognize the core elements of a reaction process: the participating molecules, their functional roles (reactant, reagent, solvent, catalyst, etc.), and associated numerical quantities (moles, mass, volume). Building on this, we design a Reaction Transformer in which each input token represents one molecule through a multi-channel composite encoding:
\begin{equation}
\begin{split}
    \mathbf{r}_j = {}&f_{\mathrm{mol}}(\mathbf{h}^{\mathrm{geo}}_j) + f_{\mathrm{amt}}(\bar{\mathbf{a}}_j) \\
                    &+ \mathbf{e}^{\mathrm{role}}(\rho_j) + \mathbf{e}^{\mathrm{type}}(\tau_j), \quad j = 1,\dots,J,
\end{split}
\end{equation}
where $\mathbf{h}^{\mathrm{geo}}_j \in \mathbb{R}^{256}$ is the GVP embedding of molecule $j$, $\bar{\mathbf{a}}_j \in \mathbb{R}^{10}$ contains normalized stoichiometric quantities and missing-value masks, $\rho_j$ is the functional role (reactant, reagent, solvent, catalyst, etc.), and $\tau_j$ distinguishes observed from masked/target molecules. Not all channels need to be present, missing values are treated as masked entries, which unifies product prediction, retrosynthesis, and yield estimation under the same architecture by simply varying which tokens are masked. For yield prediction, a dual head combines coarse classification and continuous regression, with the classification branch acting as a regularizer to stabilize the regression output. The predicted molecular embeddings and yield values are then fed back to the LLM as additional context, enabling it to continue reasoning in natural language.

\begin{figure}[t]
\centering
\includegraphics[width=\linewidth]{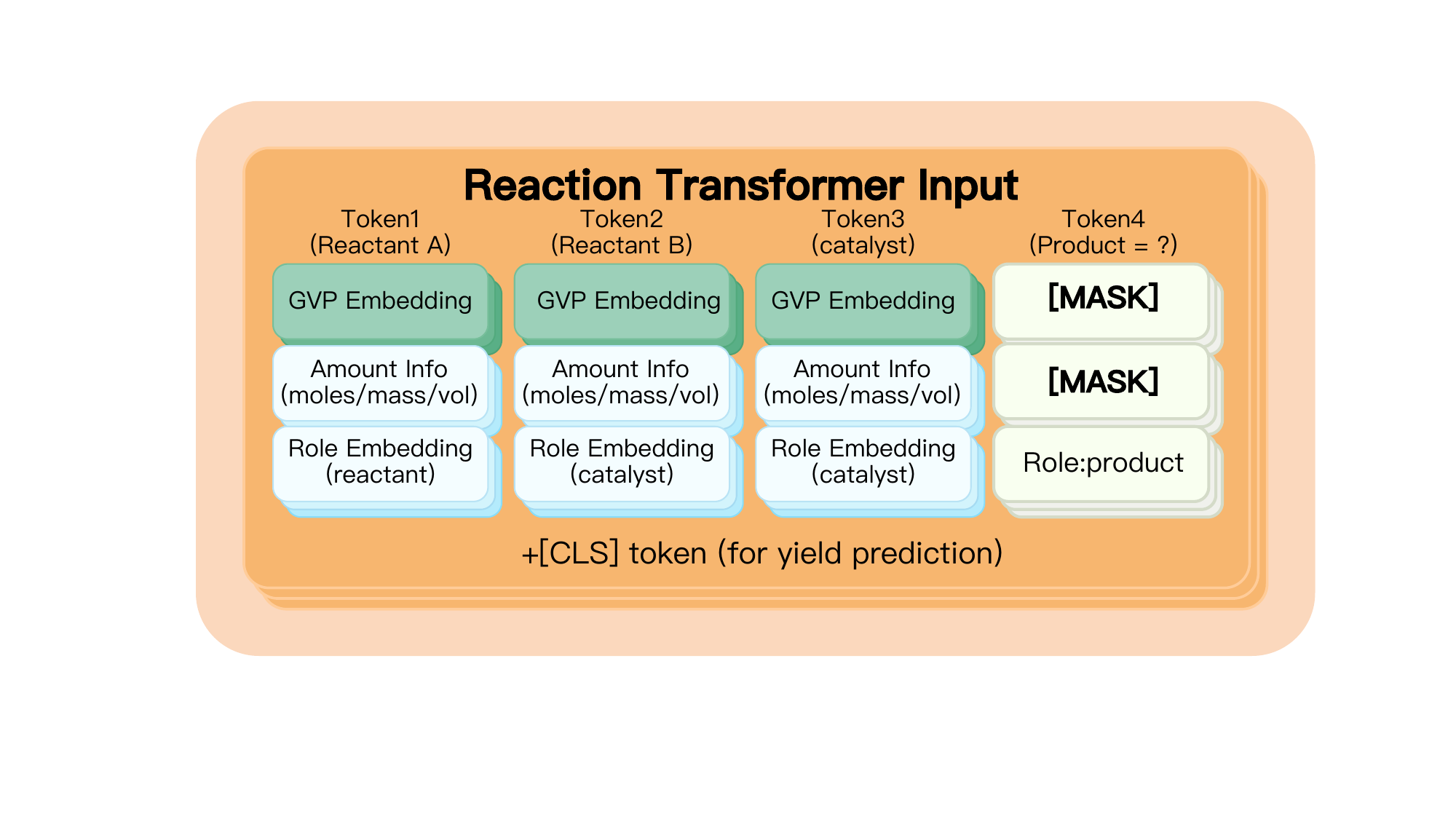}
\caption{Reaction token construction in the Reaction Sensing Module. Each token combines a GVP geometry embedding, stoichiometric amount features, and a functional role signal. Masked targets and a \texttt{[CLS]} token enable joint product and yield prediction under a unified architecture.}\label{fig:rxn_token}
\end{figure}

\subsection{Progressive Training Strategy}\label{sec:training}

\subsubsection{Stage I: Molecular Foundation Pre-training}\label{sec:stage1}

\paragraph{ \textbf{Step 1: Component Pre-training.}}
All components are pre-trained independently. For the LLM backbone, continual pre-training applies KL-divergence regularization against the frozen reference model to prevent catastrophic forgetting:
\begin{equation}
    \mathcal{L}_{\mathrm{KL}} = D_{\mathrm{KL}}\!\left(p_\theta \,\Vert\, p_{\mathrm{ref}}\right),
\end{equation}
where $p_\theta$ is the trained model's output distribution and $p_{\mathrm{ref}}$ is the frozen Qwen3-8B base. The GVP encoder is pre-trained on molecular property prediction tasks. The diffusion decoder is pre-trained on large-scale molecular databases. The Reaction Transformer is pre-trained on ORD reaction data.

\paragraph{\textbf{Step 2: Cross-Modal Alignment.}}
The LLM backbone is frozen; only the GVP encoder and MLP adapter receive gradients. For each molecule, the adapter output $\mathbf{h}^i_{\mathrm{mol}}$ and the LLM text embedding $\mathbf{h}^i_{\mathrm{text}}$ of its natural-language description form a positive pair. We apply a symmetric NT-Xent contrastive objective~\cite{simclr}:
\begin{align}
    \mathcal{L}_{\mathrm{m2t}} &= -\frac{1}{|\mathcal{B}|}\sum_{i}\log\frac{\exp(\mathrm{sim}(\mathbf{h}^i_{\mathrm{mol}},\mathbf{h}^i_{\mathrm{text}})/\tau)}{\sum_j\exp(\mathrm{sim}(\mathbf{h}^i_{\mathrm{mol}},\mathbf{h}^j_{\mathrm{text}})/\tau)}, \\
    \mathcal{L}_{\mathrm{align}} &= \tfrac{1}{2}(\mathcal{L}_{\mathrm{m2t}} + \mathcal{L}_{\mathrm{t2m}}),
\end{align}
trained on $\sim$100K SMILES-description pairs from the ChatMol corpus~\cite{chatmol} with temperature $\tau = 0.07$.

\subsubsection{\textbf{Stage 2: Joint Multi-Task Training}}\label{sec:stage2}
With the modules aligned, the full system is jointly optimized on the 300K SFT corpus. All component losses are combined as:
\begin{align}
    \mathcal{L}_{\mathrm{LM}}   &= -\textstyle\sum_t \log p_\theta(y_t \mid y_{<t}, \mathbf{X}, \tilde{\mathbf{H}}_\mathbf{X}), \\
    \mathcal{L}_{\mathrm{diff}} &= \mathbb{E}\!\left[\|\boldsymbol\epsilon - \epsilon_\theta(\mathbf{z}_t,t,\mathbf{c})\|_2^2\right], \\
    \mathcal{L}_{\mathrm{rxn}}  &= \lambda_{\mathrm{emb}}\mathcal{L}_{\mathrm{emb}} + \lambda_{\mathrm{amt}}\mathcal{L}_{\mathrm{amt}} + \lambda_{\mathrm{yield}}\mathcal{L}_{\mathrm{yield}}, \\
    \mathcal{L}_{\mathrm{yield}} &= \lambda_{\mathrm{reg}}\|\hat{y}_{\mathrm{reg}} - y\|_2^2 + \lambda_{\mathrm{cls}}\,\mathrm{CE}(\hat{\mathbf{y}}_{\mathrm{cls}},\mathrm{bin}(y)), \\
    \mathcal{L}_{\mathrm{II}}   &= \lambda_{\mathrm{LM}}\mathcal{L}_{\mathrm{LM}} + \lambda_{\mathrm{align}}\mathcal{L}_{\mathrm{align}} + \lambda_{\mathrm{diff}}\mathcal{L}_{\mathrm{diff}} + \lambda_{\mathrm{rxn}}\mathcal{L}_{\mathrm{rxn}}.
\end{align}
When a mini-batch does not contain a certain task type, the corresponding loss term is masked out.

\subsubsection{\textbf{Stage 3: Task-Specific Fine-tuning}}\label{sec:stage3}
The final stage allows specialization for particular downstream applications. Individual modules can be selectively frozen or unfrozen depending on the target task. For example, freezing the diffusion decoder and Reaction Transformer for a property-prediction-only deployment, or freezing the GVP encoder and diffusion decoder to focus training on reaction tasks. This selective fine-tuning enables practitioners to adapt the model to specific use cases without retraining the entire pipeline, while preserving the general capabilities established in Stages~1 and~2.
\section{Evaluation Details}
\label{sec:app_evaluation}

This section provides additional details for the evaluation protocol used in the main experiments. We describe the normalization procedure used to construct the capability radar chart, list the benchmark metrics included in each capability dimension, and provide representative LLM-facing prompt templates used for evaluation. Complete task-specific prompt files and parsing scripts are released with the evaluation code. Non-LLM internal module inputs, such as latent diffusion states or decoder-side latent variables, are omitted.

%--------------------------------------
\subsection{Capability Score Normalization}
\label{sec:app_normalization}
%--------------------------------------

Each capability dimension in Figure~\ref{fig:radar} aggregates one or more benchmark metrics. For each raw metric, scores are first normalized to $[0,100]$ across all evaluated models using min--max scaling:
\begin{equation}
s_i = \frac{x_i - x_{\min}}{x_{\max} - x_{\min}} \times 100,
\label{eq:normalization}
\end{equation}
where $x_i$ is the raw score of model $i$, and $x_{\min}$ and $x_{\max}$ are the minimum and maximum values among all evaluated models for the same metric. For metrics where lower values indicate better performance, such as MAE and RMSE, we first invert the metric before normalization. The final score for each capability dimension is computed as the arithmetic mean of the normalized metrics assigned to that dimension.

%--------------------------------------
\subsection{Capability Dimensions and Metrics}
\label{sec:app_metric_grouping}
%--------------------------------------

The radar chart groups benchmark metrics into five capability dimensions. Table~\ref{tab:capability_grouping} summarizes the metric grouping used in Figure~\ref{fig:radar}.

\begin{center}
\small
\captionof{table}{Metric grouping used to construct the capability radar chart.}
\label{tab:capability_grouping}
\renewcommand{\arraystretch}{1.18}
\begin{tabular}{@{}p{0.32\linewidth}p{0.60\linewidth}@{}}
\toprule
Capability dimension & Metrics included \\
\midrule
\textbf{Knowledge Core} 
& MMLU-Chemistry accuracy. \\
\addlinespace[3pt]

\textbf{Mol-Text Translation} 
& SMolInstruct name conversion metrics and molecule captioning metrics. \\
\addlinespace[3pt]

\textbf{Molecule Generation} 
& SMolInstruct molecule generation metrics and DrugR optimization metrics. \\
\addlinespace[3pt]

\textbf{Quantitative Prediction} 
& MoleculeNet regression metrics, SMolInstruct property prediction metrics, and ORD yield prediction metrics. \\
\addlinespace[3pt]

\textbf{Synthesis Reasoning} 
& SMolInstruct forward synthesis and retrosynthesis metrics, ChemBench4K product prediction and retrosynthesis metrics, and ORD reaction prediction metrics. \\
\bottomrule
\end{tabular}
\end{center}

%--------------------------------------
\subsection{Prompt Templates}
\label{sec:app_prompts}
%--------------------------------------

We demonstrate the full benchmark tables used to support the main results.

\begin{promptbox}{SMolInstruct}
{\small
For SMolInstruct, each task randomly samples one template from the corresponding \texttt{smol/<task>.json} file. The placeholder \texttt{<INPUT>} is replaced by the sample input, and residual \texttt{<OUTPUT>} placeholders are removed. When chain-of-thought prompting is disabled, we append:

\medskip
\texttt{Please only output the answer without any explanation or additional text.}

\medskip
Representative templates:
\begin{itemize}
    \item \texttt{Predict the product of a chemical reaction with \{input\} as the reactants and reagents.}
    \item \texttt{\{input\} is the IUPAC name of a molecule. Please give its SMILES representation.}
\end{itemize}
}
\end{promptbox}

\begin{promptbox}{MMLU-Chemistry}
{\small
{\ttfamily
Choose the correct option (A, B, C, or D). Return only one letter.

Question: \{question\}

Choices: A. \{choices[0]\}; B. \{choices[1]\}; C. \{choices[2]\}; D. \{choices[3]\}

Answer:
}
}
\end{promptbox}
\begin{promptbox}{ChemBench4K}
{\small
\begin{itemize}
    \item Multiple-choice: {\ttfamily
    Choose the correct option (A, B, C, or D). Return only one letter.
    
    Question: \{question\}
    
    Choices: A. \{choices[0]\}; B. \{choices[1]\}; C. \{choices[2]\}; D. \{choices[3]\}
    
    Answer:
    }
    \item Product prediction: \texttt{Predict the major product SMILES under ideal conditions. Only output the product SMILES.}
    \item Yield prediction: \texttt{Predict the reaction yield (0--100) under ideal conditions. Only output a number.}
\end{itemize}
}
\end{promptbox}

\begin{promptbox}{ORD Reaction Tasks}
{
\begin{itemize}
    \item Product prediction: \texttt{Given a reaction record with the product masked, predict the missing product SMILES. Return only the missing product SMILES. Do not output JSON or any extra text.}
    \item Reactant prediction: \texttt{Given a reaction record with one reactant masked, predict the missing reactant SMILES. Return only the missing reactant SMILES. Do not output JSON or any extra text.}
    \item Yield prediction: \texttt{Predict the isolated reaction yield as a percentage in the range [0, 100]. Return only \{"yield\_percent": float\}.}
    \item Product + yield prediction: \texttt{Given a reaction record, predict the product SMILES and the isolated reaction yield. Return only \{"products": ["SMILES", ...], "yield\_percent": float\}.}
    \item Role prediction: \texttt{Given a reaction record with a missing role/category label, predict the missing role label. Return only the label text.}
\end{itemize}

\medskip
Prompt format: \texttt{\{task\_instruction\}\\textbackslash n\\textbackslash n\{raw\_input\}}.
}
\end{promptbox}

%--------------------------------------
\subsection{Full Benchmark Tables}
\label{sec:app_full_tables}
%--------------------------------------

We include the full benchmark tables used to support the main results.

\begin{table*}[t]
\centering
\scriptsize
\setlength{\tabcolsep}{4pt}
\caption{Main results on SMolInstruct (Part I: name conversion, property prediction, and captioning).}
\label{tab:smolinstruct-main-v2}
\begin{tabular}{lccccccc}
\toprule
\textbf{Model} &
\multicolumn{2}{c}{\textbf{Name Conversion}} &
\multicolumn{4}{c}{\textbf{Property Prediction}} &
\textbf{Captioning} \\
\cmidrule(lr){2-3}\cmidrule(lr){4-7}
& I2S FTS & I2S Valid
& ESOL RMSE$\downarrow$ & Lipo RMSE$\downarrow$ & BBBP Acc & ClinTox Acc
& METEOR \\
\midrule
GPT-4o             & 55.9 & 86.7 & 5.94 & 1.34 & 52.3 & 15.3 & 11.1 \\
GPT-5              & 54.8 & 79.0 & 17.76 & 1.22 & 66.5 & 26.4 & 14.3 \\
Intern-S1-mini     & 90.7 & 76.3 & 14.58 & 1.34 & 80.2 & 80.6 & 44.6 \\
LlaSMol-Mistral-7B & 52.4 & 63.7 & 5.14 & 7.56 & 23.4 & 13.9 & 9.6 \\
Qwen3-8B Base      & 32.4 & 58.7 & 10.58 & 3.30 & 31.0 & 55.6 & 11.5 \\
Qwen3-8B-Chem      & 52.7 & 96.0 & 3.32 & 2.85 & 51.3 & 40.3 & 30.6 \\
\midrule
\textbf{Ours}      & 71.9 & 98.3 & 1.73 & 1.27 & 31.0 & 71.5 & 37.9 \\
\bottomrule
\end{tabular}
\end{table*}

\begin{table*}[t]
\centering
\scriptsize
\setlength{\tabcolsep}{6pt}
\caption{Main results on SMolInstruct (Part II: molecular generation and synthesis tasks).}
\label{tab:smolinstruct-main-v2-generation}
\begin{tabular}{lcccccc}
\toprule
\textbf{Model} &
\multicolumn{2}{c}{\textbf{Mol Generation}} &
\multicolumn{2}{c}{\textbf{Forward Synthesis}} &
\multicolumn{2}{c}{\textbf{Retrosynthesis}} \\
\cmidrule(lr){2-3}\cmidrule(lr){4-5}\cmidrule(lr){6-7}
& RDK-FTS(\%) & Valid(\%)
& RDK-FTS(\%) & Valid(\%)
& RDK-FTS(\%) & Valid(\%) \\
\midrule
GPT-4o             & 47.0 & 85.0 & 44.0 & 90.0 & 33.9 & 91.3 \\
GPT-5              & 44.9 & 79.7 & 48.3 & 89.7 & 31.1 & 88.7 \\
Intern-S1-mini     & 53.0 & 71.7 & 47.0 & 92.3 & 42.0 & 90.0 \\
LlaSMol-Mistral-7B & 16.4 & 38.7 & 15.3 & 68.7 & 8.7  & 36.7 \\
Qwen3-8B Base      & 14.2 & 47.0 & 39.6 & 81.3 & 50.7 & 89.0 \\
Qwen3-8B-Chem      & 44.5 & 93.7 & 59.4 & 98.7 & 61.1 & 98.7 \\
\midrule
\textbf{Ours}      & 61.3 & 93.7 & 70.0 & 97.7 & 63.7 & 99.3 \\
\bottomrule
\end{tabular}
\end{table*}

\begin{table*}[t]
\centering
\small
\setlength{\tabcolsep}{7pt}
\caption{Main results on MMLU (Acc\%).}
\label{tab:main-mmlu-subsets}
\begin{tabular}{lccccc}
\toprule
\textbf{Model} &
\textbf{HS Chem} &
\textbf{College Chem} &
\textbf{Organic Chem} &
\textbf{Physical Chem} &
\textbf{General Sci} \\
\midrule
GPT-4o                & 76.85 & 54.00 & 70.63 & 77.46 & 80.78 \\
GPT-5                 & 82.75 & 64.00 & 72.93 & 80.73 & 83.12 \\
Intern-S1-mini        & 76.85 & 57.00 & 70.30 & 71.52 & 75.63 \\
LlaSMol-Mistral-7B    & 52.67 & 46.68 & 66.67 & 65.84 & 71.54 \\
Qwen3-8B              & 67.98 & 57.00 & 64.36 & 71.31 & 75.16 \\
Qwen3-8B-Chem         & 70.94 & 57.00 & 66.34 & 70.49 & 75.63 \\
\midrule
\textbf{Ours}         & 72.41 & 57.00 & 67.33 & 68.65 & 72.97 \\
\bottomrule
\end{tabular}
\end{table*}

\begin{table*}[t]
\centering
\small
\setlength{\tabcolsep}{7pt}
\caption{Main results on ChemBench4K (Acc\%).}
\label{tab:chembench4k-main}
\begin{tabular}{lcccccc}
\toprule
\textbf{Model} &
\textbf{Product} &
\textbf{Retro} &
\textbf{Yield} &
\textbf{Cap2Mol} &
\textbf{Mol2Cap} &
\textbf{Average} \\
\midrule
GPT-4o                & 91.00 & 58.00 & 45.66 & 96.32 & 60.25 & 70.25 \\
GPT-5                 & 93.66 & 78.00 & 43.00 & 99.33 & 64.62 & 75.72 \\
Intern-S1-mini        & 22.00 & 24.00 & 30.00 & 22.41 & 22.25 & 24.13 \\
LlaSMol-Mistral-7B    & 22.33 & 23.33 & 32.33 & 35.79 & 21.88 & 27.13 \\
Qwen3-8B              & 22.00 & 24.00 & 30.00 & 22.25 & 22.41 & 24.13 \\
Qwen3-8B-Chem         & 51.00 & 37.67 & 32.67 & 80.27 & 36.88 & 47.70 \\
\midrule
\textbf{Ours}         & 92.00 & 80.67 & 38.67 & 94.31 & 49.62 & 71.05 \\
\bottomrule
\end{tabular}
\end{table*}

\begin{table*}[t]
\centering
\scriptsize
\setlength{\tabcolsep}{4pt}
\caption{Main results on the ORD testset.}
\label{tab:ord-main}
\begin{tabular}{lccccccc ccc cc}
\toprule
\multirow{2}{*}{\textbf{Model}} &
\multicolumn{2}{c}{\textbf{Product Prediction}} &
\multicolumn{5}{c}{\textbf{Product + Yield Prediction}} &
\multicolumn{3}{c}{\textbf{Yield Prediction}} &
\multicolumn{2}{c}{\textbf{Reactant Prediction}} \\
\cmidrule(lr){2-3}
\cmidrule(lr){4-8}
\cmidrule(lr){9-11}
\cmidrule(lr){12-13}
& Valid & RDK-FTS
& Valid & RDK-FTS & MAE$\downarrow$ & Acc($\pm1$) & NDCG
& MAE$\downarrow$ & Acc($\pm1$) & NDCG
& Valid & RDK-FTS \\
\midrule
GPT-4o                & 57.0  & 55.10 & 91.0  & 54.72 & 0.36  & 29.6  & 33.7  & 0.29  & 33.7  & 38.2  & 96.1  & 77.40 \\
GPT-5                 & 82.0  & 72.73 & 92.1  & 74.27 & 0.35  & 29.6  & 35.8  & 0.32  & 29.6  & 36.0  & 97.0  & 89.21 \\
Intern-S1-mini        & 67.3  & 49.60 & 100.0 & 50.80 & 0.50  & 19.2  & 31.4  & 0.54  & 15.0  & 28.2  & 83.3  & 63.57 \\
LlaSMol-Mistral-7B    & 83.0  & 25.10 & 55.0  & 23.80 & 0.49  & 18.5  & 31.7  & 0.53  & 17.7  & 29.8  & 57.3  & 24.23 \\
Qwen3-8B              & 45.56 & 14.40 & 16.36 & 20.15 & 0.49  & 20.30 & 31.39 & 0.33  & 33.62 & 38.81 & 43.62 & 13.84 \\
Ours w/o Reaction Module       & 11.38 & 16.65 & 11.51 & 17.87 & 0.52  & 18.48 & 29.60 & 0.27  & 33.33 & 39.36 & 24.64 & 9.18  \\
\midrule
\textbf{Ours}         & 42.0  & 21.67 & 97.6  & 22.43 & 0.343 & 31.2  & 35.5  & 0.325 & 32.9  & 39.4  & 74.7  & 15.91 \\
\bottomrule
\end{tabular}
\end{table*}

\newpage{}
\section{Acknowledgments}

This work is partially supported by Tsinghua University (Department of Computer Science and Technology) - Sinopec Joint Research Center for Artificial Intelligence.

\newpage
\bibliographystyle{plainnat}
\bibliography{references}

\end{document}